\documentclass{article} % For LaTeX2e
\usepackage{iclr2025_conference,times}
\usepackage[utf8]{inputenc}
\usepackage{graphicx}
\usepackage{amssymb}
\usepackage{xcolor}
\usepackage{multirow}
\usepackage{placeins}
\usepackage{float} 

\usepackage{amsmath,amsfonts,bm}

\def\eqref#1{equation~\ref{#1}}
\def\1{\bm{1}}

\DeclareMathAlphabet{\mathsfit}{\encodingdefault}{\sfdefault}{m}{sl}
\SetMathAlphabet{\mathsfit}{bold}{\encodingdefault}{\sfdefault}{bx}{n}

\usepackage{tabularx} % ⬅️ 자동 줄바꿈을 위한 패키지
\usepackage{array} % ⬅️ 테이블 정렬을 위한 패키지
\usepackage{hyperref}
\usepackage{url}
\title{TWICE: what advanTages can loW-resource domaIn-speCific Embedding model bring?
— A Case Study on Korea Financial Texts}

\author{Yewon Hwang\thanks{These authors contributed equally.} \\
MODULABS, Financial NLP Lab \\
\texttt{yeowonh@sju.ac.kr}
\And
Sungbum Jung\footnotemark[1] \\
MODULABS, Financial NLP Lab \\
\texttt{jsbreset@gmail.com}
\And
Sara Yu\footnotemark[1] \\
KT Corporation \\
\texttt{sara.yu@kt.com}
\And
Hanwool Lee\footnotemark[1]\hspace{0.5em}\thanks{Corresponding author} \\
Shinhan Securities Co, MODULABS, Financial NLP Lab \\
\texttt{gksdnf424@gmail.com}
}

\iclrfinalcopy % Uncomment for camera-ready version, but NOT for submission.
\begin{document}

\maketitle

\begin{abstract}
Domain specificity of embedding models is critical for the effective performance. However, existing benchmarks, such as FinMTEB, are primarily designed for high-resource languages, leaving low-resource settings, such as Korean, under-explored. Directly translating established English benchmarks often fails to capture the linguistic and cultural nuances present in low-resource domains. In this paper, titled \textbf{TWICE}: what advan\textbf{T}ages can lo\textbf{W}-resource doma\textbf{I}n-specifi\textbf{C} \textbf{E}mbedding model bring?— A Case Study on Korea Financial Texts, we introduce \textbf{KorFinMTEB}, a novel benchmark for the Korean financial domain, specifically tailored to reflect its unique cultural characteristics in low-resources languages. Our experimental results reveal that while the models perform robustly on a translated version of FinMTEB, their performance on KorFinMTEB uncovers subtle yet critical discrepancies—especially in tasks requiring deeper semantic understanding—that underscore the limitations of direct translation. This discrepancy underscores the limitations of direct translation and highlights the necessity of benchmarks that incorporate language-specific idiosyncrasies and cultural nuances. The insights from our study advocate for the development of domain-specific evaluation frameworks that can more accurately assess and drive the progress of embedding models in low-resource settings.
\end{abstract}

\section{Introduction}

Embedding models have revolutionized NLP, with benchmarks such as MTEB \citep{muennighoff2023mtebmassivetextembedding} and FinMTEB \citep{tang2024needdomainspecificembeddingmodels} providing robust evaluations for high-resource languages and specialized domains like finance. However, low-resource languages like Korean are underrepresented. Directly translating existing benchmarks often introduces context loss and fails to capture cultural nuances \citep{son2024kmmlumeasuringmassivemultitask, son2024haeraebenchevaluationkorean}—a critical issue in financial texts where precise terminology is paramount \citep{wu2023bloomberggptlargelanguagemodel}.

To address these challenges, we propose \textbf{KorFinMTEB}, a novel benchmark built from authentic Korean financial texts as shown in Figure 1. It reflects the unique linguistic and cultural characteristics of Korea's financial domain. Our comparative analysis between a directly translated version of FinMTEB and KorFinMTEB reveals a significant performance gap, underscoring the inadequacy of simple translation for evaluating domain-specific models in low-resource settings.

Our contributions are threefold:
\begin{enumerate}
    \item We demonstrate the limitations of directly translated benchmarks in capturing low-resource language nuances and culture.
    \item We introduce KorFinMTEB, a benchmark consisting of 7 tasks with 26 datasets tailored for Korean financial texts.
    \item We provide a comparative analysis that highlights the need for customized evaluation frameworks of Korean Financial Domain-Specific Task.
\end{enumerate}

The dataset of KorFinMTEB benchmark are fully open-sourced and publicly available, ensuring reproducibility and transparency.

\begin{figure}[h]
\begin{center}
\includegraphics[width=0.7\linewidth]{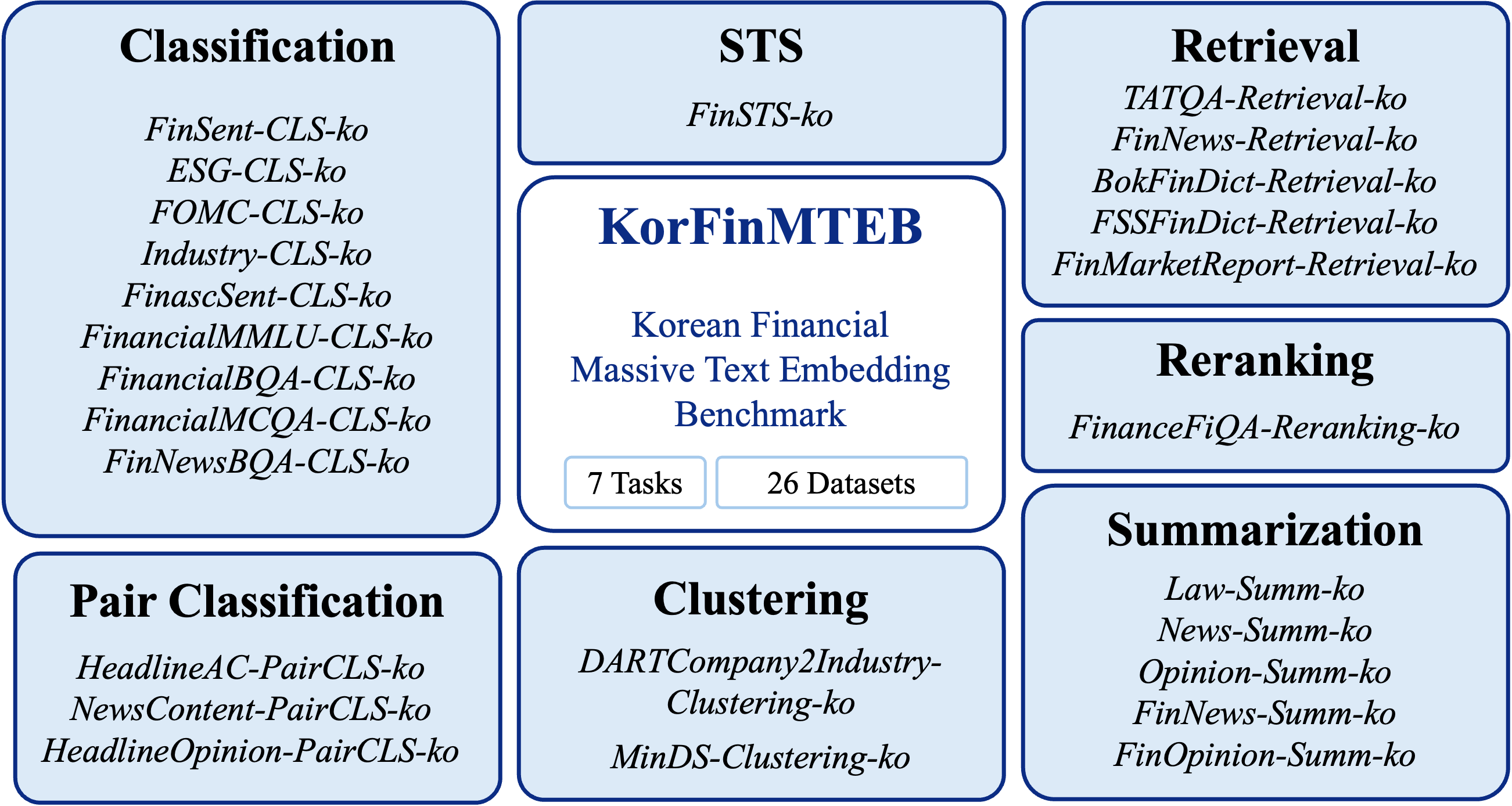} 
\end{center}
\caption{An overview of 7 tasks and 26 datasets used in KorFinMTEB.}
\end{figure}

\section{Related Works}
Recent advances in neural embedding models have evolved from early distributed word representations such as word2vec \citep{mikolov2013efficientestimationwordrepresentations} to contextual approaches like BERT \citep{devlin2019bertpretrainingdeepbidirectional}. Subsequent methods, including SentenceBERT \citep{reimers2019sentencebertsentenceembeddingsusing} and SimCSE \citep{gao2022simcsesimplecontrastivelearning}, further enhanced semantic representation, while multilingual frameworks like InfoXLM \citep{chi2021infoxlminformationtheoreticframeworkcrosslingual} extend these benefits to low-resource languages.

Despite these improvements, general-purpose models often fall short in domain-specific applications. For instance, in finance and biomedicine, tailored models such as FinBERT \citep{araci2019finbertfinancialsentimentanalysis} and BioBERT \citep{Lee_2019} capture specialized terminology and nuances that generic embeddings may overlook. Recent financial NLP studies even report that models like SentenceBERT and Ada embeddings tend to overestimate similarity in reports with minor surface variations \citep{liu2024surfacesimilaritydetectingsubtle}, highlighting the necessity for domain-specific benchmarks like FinMTEB \citep{tang2024needdomainspecificembeddingmodels} that can provide more accurate evaluations in specialized contexts.

\section{Benchmark Construction and Experimental Setup}
\subsection{Benchmark Construction}
In designing KorFinMTEB, we adopt the FinMTEB format and extend it to the Korean financial domain, by incorporating seven core tasks: classification, clustering, retrieval, summarization, pair classification, reranking, and semantic textual similarity. For each task, we integrate openly available datasets with in-house curated data to comprehensively cover the linguistic and domain-specific challenges inherent in financial text analysis.

\paragraph{Classification} We define nine classification subtasks to capture diverse financial phenomena. For example, \emph{FinancialNews-CLS} categorizes financial news as Positive, Negative, or Neutral, while \emph{ESG-CLS-ko} filters ESG-related news into E/S/G/Non-ESG classes. Additional subtasks include KorFOMCClassification (distinguishing Hawkish/Dovish/Neutral tones based on key financial terms), IndustryClassification (assigning industry labels to analytical reports), and several variants of sentiment and QA classifications (e.g., \texttt{FinSent-CLS-ko}, 
\texttt{FinancialMMLU-CLS-ko}, \texttt{FinancialBQA-CLS-ko}, \texttt{FinancialMCQA-CLS-ko}, and \texttt{FinNewsBQA-CLS-ko}). Data sources include the \textit{korfin-asc} dataset \citep{son2023removingnonstationaryknowledgepretrained}, Hugging Face datasets (e.g., \texttt{allganize/financial-mmlu-ko} and FINNUMBER/QA\_Instruction), and AI Hub’s financial news reading comprehension data. Where open data was insufficient, we constructed additional datasets to ensure task completeness.

\paragraph{Retrieval} The retrieval component is built to reflect the complexity of financial queries that often involve both textual and tabular information. We include tasks based on datasets such as \texttt{allganize/flare-convfinqa-multiturn-ko} and its subset \texttt{allganize/flare-convfinqa-ko}, supplemented by AI Hub’s news article comprehension data. Additional retrieval tasks (e.g., \texttt{BokFinDict}, \texttt{FSSFinDict}, \texttt{TATQA}, \texttt{FinNews}, and \texttt{FinMarketReport}) were developed to further challenge the models’ ability to fetch domain-relevant information.

\paragraph{Clustering, Summarization, Pair Classification, Reranking, and Semantic Similarity} For clustering, we employ datasets such as crawled DART (Korean corporate disclosure) data, along with AI Hub petitions Q\&A records, to assess the models’ capacity to group similar financial entities or documents. Summarization tasks, including \texttt{Law-Summ-ko}, \texttt{News-Summ-ko}, \texttt{Opinion-Summ-ko}, \texttt{FinNews-Summ-ko}, and \texttt{FinOpinion-Summ-ko}, leverage AI Hub’s document summarization texts. For pair classification, which examines semantic relationships between text pairs, we combine data from \textit{korfin\-asc} and \texttt{Sujet-Finance-Instruct-177k-ko} . The reranking task (\texttt{FinanceFiQA-Reranking-ko}) is based on the BCCard Finance QnA dataset, and semantic textual similarity is evaluated using \texttt{FinSTS-ko} the only pre-existing financial embedding benchmark adopted after quality verification. The quality verification was carried out by people with specialized knowledge in the financial sector, such as masters in economics and employees of financial firms.

\paragraph{Overall Approach} By adhering to a modular design that mirrors FinMTEB’s structure, KorFinMTEB provides a cohesive yet challenging evaluation suite tailored to the nuances of the Korean financial domain. Our deliberate combination of openly available resources with bespoke data curation ensures that each task reflects real-world financial text complexities, thereby facilitating robust and domain-specific model assessments.

\subsection{Experimental Setup}
To assess the impact of low-resource domain-specific data, we compare the performance of embedding models on two benchmark variants: (1) \textbf{Trans-ko-FinMTEB}, a version of FinMTEB translated into Korean using GPT-4o, and (2) \textbf{KorFinMTEB}, our newly constructed benchmark based on native Korean financial data. Our primary hypothesis is that performance differences will emerge between the two datasets, highlighting the benefits of using authentic, domain-specific data over simple synthetic translations.

For the experiments, we selected a suite of state-of-the-art embedding models that have achieved top-tier performance on the MTEB leaderboard\citep{muennighoff2023mtebmassivetextembedding} or are widely adopted in the community:
\begin{itemize}
    \item \textbf{bge-en-icl} \citep{li2024makingtextembeddersfewshot}
    \item \textbf{gte-Qwen2-1.5B-instruct} \citep{li2023towards}
    \item \textbf{e5-mistral-7b-instruct} \citep{wang2023improving}
    \item \textbf{bge-large-en-v1.5} \citep{li2024makingtextembeddersfewshot}
    \item \textbf{text-embedding-3-small} (OpenAI; see \url{https://openai.com/index/new-embedding-models-and-api-updates/})
    \item \textbf{instructor-base} \citep{su2023embeddertaskinstructionfinetunedtext}
    \item \textbf{all-MiniLM-L12-v2} \citep{wang2020minilmdeepselfattentiondistillation}
\end{itemize}

Additionally, to verify whether an embedding model trained in Korean performs better on our benchmark, KorFinMTEB, we used KURE-v1, a BGE-M3 based model fine-tuned in Korean, for comparative experiments.

\begin{itemize}
    \item \textbf{kure-v1}\citep{KURE}
\end{itemize}

We conduct evaluations across a range of tasks—including classification, clustering, retrieval, summarization, pair classification, reranking, and semantic textual similarity—using metrics as defined in the FinMTEB framework (e.g., accuracy for classification, etc.). All models are evaluated under a consistent set of hyperparameters and experimental configurations across both benchmark variants to isolate the effect of the dataset's linguistic and cultural nuances.

By contrasting model performance on TranslatedFinMTEB and our proposed KorFinMTEB, experiments aim to quantify the advantages of employing native-domain benchmarks in low-resource settings. Further implementation details, including dataset preprocessing and task-specific configurations, are provided in the supplementary material.

The experimental results for the above models across specific tasks are shown in the Table \ref{tab:two-col-diff}.

\subsection{Results and Analysis}
\begin{table}[t]
\centering
\begin{minipage}[t]{0.48\textwidth}
\centering
\footnotesize
\renewcommand{\arraystretch}{1.0}
\begin{tabular}{l c}
\hline
\multicolumn{2}{c}{\textbf{FOMC Classification}} \\
\hline
bge-en-icl              & \textcolor{red}{$\blacktriangle$} 0.180  \\
gte-Qwen2-1.5B-instruct & \textcolor{blue}{$\blacktriangledown$} -0.160 \\
e5-mistral-7b-instruct  & \textcolor{red}{$\blacktriangle$} 0.165 \\
bge-large-en-v1.5       & \textcolor{blue}{$\blacktriangledown$} -0.195\\
text-embedding-3-small  & \textcolor{red}{$\blacktriangle$} 0.150  \\
all-MiniLM-L12-v2       & \textcolor{blue}{$\blacktriangledown$} -0.160 \\
instructor-base         & \textcolor{blue}{$\blacktriangledown$} -0.190 \\
kure-v1                 & \textcolor{blue}{$\blacktriangledown$} -0.140 \\
\hline
\multicolumn{2}{c}{\textbf{ESG Classification}} \\
\hline
bge-en-icl              & \textcolor{blue}{$\blacktriangledown$} -0.130 \\
gte-Qwen2-1.5B-instruct & \textcolor{red}{$\blacktriangle$} 0.160   \\
e5-mistral-7b-instruct  & \textcolor{blue}{$\blacktriangledown$} -0.155 \\
bge-large-en-v1.5       & \textcolor{red}{$\blacktriangle$} 0.080   \\
text-embedding-3-small  & \textcolor{red}{$\blacktriangle$} 0.100   \\
all-MiniLM-L12-v2       & \textcolor{red}{$\blacktriangle$} 0.015  \\
instructor-base         & \textcolor{blue}{$\blacktriangledown$} -0.055 \\
kure-v1                 & \textcolor{red}{$\blacktriangle$} 0.160 \\
\hline
\multicolumn{2}{c}{\textbf{FinNews Classification}} \\
\hline
bge-en-icl              & \textcolor{red}{$\blacktriangle$} 0.045 \\
gte-Qwen2-1.5B-instruct & \textcolor{red}{$\blacktriangle$} 0.060  \\
e5-mistral-7b-instruct  & \textcolor{red}{$\blacktriangle$} 0.030  \\
bge-large-en-v1.5       & \textcolor{red}{$\blacktriangle$} 0.065 \\
text-embedding-3-small  & \textcolor{red}{$\blacktriangle$} 0.095 \\
all-MiniLM-L12-v2       & \textcolor{red}{$\blacktriangle$} 0.100 \\
instructor-base         & \textcolor{red}{$\blacktriangle$} 0.065 \\
kure-v1                 & \textcolor{blue}{$\blacktriangledown$} -0.005 \\
\hline
\multicolumn{2}{c}{\textbf{Semantic Textual Similarity}} \\
\hline
bge-en-icl              & \textcolor{red}{$\blacktriangle$} 0.096 \\
gte-Qwen2-1.5B-instruct & \textcolor{red}{$\blacktriangle$} 0.180 \\
e5-mistral-7b-instruct  & \textcolor{red}{$\blacktriangle$} 0.116 \\
bge-large-en-v1.5       & \textcolor{red}{$\blacktriangle$} 0.089 \\
text-embedding-3-small  & \textcolor{red}{$\blacktriangle$} 0.183 \\
all-MiniLM-L12-v2       & \textcolor{red}{$\blacktriangle$} 0.054 \\
instructor-base         & \textcolor{blue}{$\blacktriangledown$} -0.073 \\
kure-v1                 & \textcolor{red}{$\blacktriangle$} 0.147 \\
\hline
\multicolumn{2}{c}{\textbf{PairClassification}} \\
\hline
bge-en-icl              & \textcolor{red}{$\blacktriangle$} 0.325 \\
gte-Qwen2-1.5B-instruct & \textcolor{red}{$\blacktriangle$} 0.334 \\
e5-mistral-7b-instruct  & \textcolor{red}{$\blacktriangle$} 0.335 \\
bge-large-en-v1.5       & \textcolor{red}{$\blacktriangle$} 0.314 \\
text-embedding-3-small  & \textcolor{red}{$\blacktriangle$} 0.321 \\
all-MiniLM-L12-v2       & \textcolor{red}{$\blacktriangle$} 0.266 \\
instructor-base         & \textcolor{red}{$\blacktriangle$} 0.455 \\
kure-v1                 & \textcolor{red}{$\blacktriangle$} 0.086 \\
\hline
\end{tabular}
\end{minipage}
\hfill
\begin{minipage}[t]{0.48\textwidth}
\centering
\footnotesize
\renewcommand{\arraystretch}{1.0}
\begin{tabular}{l c}
\hline
\multicolumn{2}{c}{\textbf{TAT QA Retrieval}} \\
\hline
bge-en-icl              & \textcolor{red}{$\blacktriangle$} 0.108 \\
gte-Qwen2-1.5B-instruct & \textcolor{blue}{$\blacktriangledown$} -0.395 \\
e5-mistral-7b-instruct  & \textcolor{blue}{$\blacktriangledown$} -0.462 \\
bge-large-en-v1.5       & \textcolor{red}{$\blacktriangle$} 0.072 \\
text-embedding-3-small  & \textcolor{blue}{$\blacktriangledown$} -0.520 \\
all-MiniLM-L12-v2       & \textcolor{red}{$\blacktriangle$} 0.080 \\
instructor-base         & \textcolor{red}{$\blacktriangle$} 0.197 \\
kure-v1                 & \textcolor{blue}{$\blacktriangledown$} -0.693 \\
\hline
\multicolumn{2}{c}{\textbf{GoldmanEncRetrieval (vs. FssDict)}} \\
\hline
bge-en-icl              & \textcolor{blue}{$\blacktriangledown$} -0.100 \\
gte-Qwen2-1.5B-instruct & \textcolor{blue}{$\blacktriangledown$} -0.273 \\
e5-mistral-7b-instruct  & \textcolor{blue}{$\blacktriangledown$} -0.260 \\
bge-large-en-v1.5       & \textcolor{blue}{$\blacktriangledown$} -0.218 \\
text-embedding-3-small  & \textcolor{blue}{$\blacktriangledown$} -0.150 \\
all-MiniLM-L12-v2       & \textcolor{blue}{$\blacktriangledown$} -0.078 \\
instructor-base         & \textcolor{blue}{$\blacktriangledown$} -0.016 \\
kure-v1                 & \textcolor{blue}{$\blacktriangledown$} -0.272 \\
\hline
\multicolumn{2}{c}{\textbf{GoldmanEncRetrieval (vs. BokDict)}} \\
\hline
bge-en-icl              & \textcolor{red}{$\blacktriangle$} 0.060 \\
gte-Qwen2-1.5B-instruct & \textcolor{blue}{$\blacktriangledown$} -0.296 \\
e5-mistral-7b-instruct  & \textcolor{blue}{$\blacktriangledown$} -0.289 \\
bge-large-en-v1.5       & \textcolor{blue}{$\blacktriangledown$} -0.218 \\
text-embedding-3-small  & \textcolor{blue}{$\blacktriangledown$} -0.148 \\
all-MiniLM-L12-v2       & \textcolor{blue}{$\blacktriangledown$} -0.008 \\
instructor-base         & \textcolor{blue}{$\blacktriangledown$} -0.026 \\
kure-v1                 & \textcolor{blue}{$\blacktriangledown$} -0.248 \\
\hline
\multicolumn{2}{c}{\textbf{Reranking}} \\
\hline
bge-en-icl              & \textcolor{red}{$\blacktriangle$} 0.525 \\
gte-Qwen2-1.5B-instruct & \textcolor{red}{$\blacktriangle$} 0.598 \\
e5-mistral-7b-instruct  & \textcolor{red}{$\blacktriangle$} 0.368 \\
bge-large-en-v1.5       & \textcolor{red}{$\blacktriangle$} 0.314 \\
text-embedding-3-small  & \textcolor{red}{$\blacktriangle$} 0.320 \\
all-MiniLM-L12-v2       & \textcolor{red}{$\blacktriangle$} 0.266 \\
instructor-base         & \textcolor{red}{$\blacktriangle$} 0.455 \\
kure-v1                 & \textcolor{red}{$\blacktriangle$} 0.086 \\
\hline
\multicolumn{2}{c}{\textbf{Clustering}} \\
\hline
bge-en-icl              & \textcolor{red}{$\blacktriangle$} 0.477 \\
gte-Qwen2-1.5B-instruct & \textcolor{red}{$\blacktriangle$} 0.384 \\
e5-mistral-7b-instruct  & \textcolor{red}{$\blacktriangle$} 0.361 \\
bge-large-en-v1.5       & \textcolor{red}{$\blacktriangle$} 0.314 \\
text-embedding-3-small  & \textcolor{red}{$\blacktriangle$} 0.414 \\
all-MiniLM-L12-v2       & \textcolor{red}{$\blacktriangle$} 0.036 \\
instructor-base         & \textcolor{red}{$\blacktriangle$} 0.049 \\
kure-v1                 & \textcolor{red}{$\blacktriangle$} 0.321 \\
\hline

\end{tabular}
\end{minipage}
\caption{Differences \(\text{(FinMTEB} - \text{KorFinMTEB)}\) across tasks and models. Red \textcolor{red}{$\blacktriangle$} indicates a positive difference; 
blue \textcolor{blue}{$\blacktriangledown$} indicates a negative difference. 
}
\label{tab:two-col-diff}
\end{table}

We evaluated renowned embedding models on two benchmarks: \textbf{KorFinMTEB}, built from native Korean financial texts, and \textbf{Translated-FinMTEB}, obtained by directly translating FinMTEB via GPT-4o. Although both benchmarks adhere to identical task formats, their performance distributions differ markedly, revealing that translation-based approaches fail to capture the full range of domain-specific nuances present in authentic Korean texts.

For straightforward classification tasks (e.g., FinSent-CLS-ko), the performance differences were modest because core financial terminology is largely preserved during translation; in some cases, Translated-FinMTEB even achieved marginally higher accuracy owing to the reduced linguistic variability. However, for tasks demanding deeper semantic understanding—such as semantic textual similarity, pair classification, and summarization—models consistently exhibited a 5–8\% performance drop on KorFinMTEB. These findings strongly suggest that native data encapsulates richer linguistic subtleties and culturally embedded domain expressions that are diluted in translated texts, thereby highlighting the inherent limitations of a translation-based benchmark.

Intriguingly, retrieval tasks (e.g., \emph{TATQA-Retrieval-ko} and \emph{FinNews-Retrieval-ko}) proved more challenging on Translated-FinMTEB. Analysis indicates that translation artifacts often generate unnatural or ambiguous query phrasing, leading to a misalignment with corpus entries that retain genuine Korean terminology and context. This paradox not only alters task difficulty but also reinforces the necessity for benchmarks constructed from native sources. Notably, models fine-tuned on Korean data (e.g., \textbf{kure-v1}) demonstrated more robust performance, emphasizing the benefits of language-specific training for domain-specific applications.

In summary, our results provide compelling evidence that \textbf{KorFinMTEB} more faithfully reflects the complexities of Korean financial discourse and serves as a more reliable evaluation framework for embedding models in low-resource settings. These insights advocate for the adoption of native, domain-specific benchmarks to drive model improvements and ensure real-world applicability.

\section{Limitations}

Although \textbf{KorFinMTEB} covers a diverse range of tasks and sub-domains within Korean finance, several limitations remain. First, certain niche financial topics (e.g., official financial analyst reports) are underrepresented due to limited publicly available datasets. Second, we focus primarily on text-based tasks, leaving related modalities (e.g., tables with rich numerical information) for future extensions. Third, while we tested various state-of-the-art embedding models, our exploration of hyperparameter tuning and advanced optimization techniques was constrained by computational resources and paper length. Finally, our benchmark primarily evaluates sentence- or paragraph-level embeddings and does not fully capture document-level reasoning, which is often critical in real-world financial decision-making. Addressing these issues will require ongoing collaboration among researchers, practitioners, and data providers to expand both the scope and granularity of the benchmark.

\section{Conclusion}

In this paper, we introduced \textbf{KorFinMTEB}, a benchmark composed of native Korean financial texts, alongside experimental results on both KorFinMTEB and a translation-based counterpart. Our findings reveal that translated benchmarks often fail to reflect the linguistic and contextual depth of low-resource domains, leading to inflated or inconsistent performance metrics. By contrast, KorFinMTEB provides a more authentic and robust evaluation framework, spotlighting the importance of domain-specific expressions and cultural nuances. 

Moreover, models fine-tuned on Korean data achieve more stable outcomes across tasks, indicating that in-language training is essential for capturing the intricacies of specialized domains like finance. We believe that developing similar native benchmarks for other low-resource languages will enhance the overall reliability and applicability of embedding models in real-world, domain-rich contexts.

% \subsubsection*{Author Contributions}
% Camera-Ready Version때 작성
% If you'd like to, you may include  a section for author contributions as is done
% in many journals. This is optional and at the discretion of the authors.

\bibliography{iclr2025_conference}

\begin{thebibliography}{20}
\providecommand{\natexlab}[1]{#1}
\providecommand{\url}[1]{\texttt{#1}}
\expandafter\ifx\csname urlstyle\endcsname\relax
  \providecommand{\doi}[1]{doi: #1}\else
  \providecommand{\doi}{doi: \begingroup \urlstyle{rm}\Url}\fi

\bibitem[Araci(2019)]{araci2019finbertfinancialsentimentanalysis}
Dogu Araci.
\newblock Finbert: Financial sentiment analysis with pre-trained language models, 2019.
\newblock URL \url{https://arxiv.org/abs/1908.10063}.

\bibitem[Chi et~al.(2021)Chi, Dong, Wei, Yang, Singhal, Wang, Song, Mao, Huang, and Zhou]{chi2021infoxlminformationtheoreticframeworkcrosslingual}
Zewen Chi, Li~Dong, Furu Wei, Nan Yang, Saksham Singhal, Wenhui Wang, Xia Song, Xian-Ling Mao, Heyan Huang, and Ming Zhou.
\newblock Infoxlm: An information-theoretic framework for cross-lingual language model pre-training, 2021.
\newblock URL \url{https://arxiv.org/abs/2007.07834}.

\bibitem[Devlin et~al.(2019)Devlin, Chang, Lee, and Toutanova]{devlin2019bertpretrainingdeepbidirectional}
Jacob Devlin, Ming-Wei Chang, Kenton Lee, and Kristina Toutanova.
\newblock Bert: Pre-training of deep bidirectional transformers for language understanding, 2019.
\newblock URL \url{https://arxiv.org/abs/1810.04805}.

\bibitem[Gao et~al.(2022)Gao, Yao, and Chen]{gao2022simcsesimplecontrastivelearning}
Tianyu Gao, Xingcheng Yao, and Danqi Chen.
\newblock Simcse: Simple contrastive learning of sentence embeddings, 2022.
\newblock URL \url{https://arxiv.org/abs/2104.08821}.

\bibitem[Jang et~al.(2024)Jang, Son, and Lee]{KURE}
Youngjoon Jang, Junyoung Son, and Taemin Lee, 2024.
\newblock URL \url{https://github.com/nlpai-lab/KURE}.

\bibitem[Lee et~al.(2019)Lee, Yoon, Kim, Kim, Kim, So, and Kang]{Lee_2019}
Jinhyuk Lee, Wonjin Yoon, Sungdong Kim, Donghyeon Kim, Sunkyu Kim, Chan~Ho So, and Jaewoo Kang.
\newblock Biobert: a pre-trained biomedical language representation model for biomedical text mining.
\newblock \emph{Bioinformatics}, 36\penalty0 (4):\penalty0 1234–1240, September 2019.
\newblock ISSN 1367-4811.
\newblock \doi{10.1093/bioinformatics/btz682}.
\newblock URL \url{http://dx.doi.org/10.1093/bioinformatics/btz682}.

\bibitem[Li et~al.(2024)Li, Qin, Xiao, Chen, Luo, Shao, Lian, and Liu]{li2024makingtextembeddersfewshot}
Chaofan Li, MingHao Qin, Shitao Xiao, Jianlyu Chen, Kun Luo, Yingxia Shao, Defu Lian, and Zheng Liu.
\newblock Making text embedders few-shot learners, 2024.
\newblock URL \url{https://arxiv.org/abs/2409.15700}.

\bibitem[Li et~al.(2023)Li, Zhang, Zhang, Long, Xie, and Zhang]{li2023towards}
Zehan Li, Xin Zhang, Yanzhao Zhang, Dingkun Long, Pengjun Xie, and Meishan Zhang.
\newblock Towards general text embeddings with multi-stage contrastive learning.
\newblock \emph{arXiv preprint arXiv:2308.03281}, 2023.

\bibitem[Liu et~al.(2024)Liu, Yang, and Tam]{liu2024surfacesimilaritydetectingsubtle}
Jiaxin Liu, Yi~Yang, and Kar~Yan Tam.
\newblock Beyond surface similarity: Detecting subtle semantic shifts in financial narratives, 2024.
\newblock URL \url{https://arxiv.org/abs/2403.14341}.

\bibitem[Mikolov et~al.(2013)Mikolov, Chen, Corrado, and Dean]{mikolov2013efficientestimationwordrepresentations}
Tomas Mikolov, Kai Chen, Greg Corrado, and Jeffrey Dean.
\newblock Efficient estimation of word representations in vector space, 2013.
\newblock URL \url{https://arxiv.org/abs/1301.3781}.

\bibitem[Muennighoff et~al.(2023)Muennighoff, Tazi, Magne, and Reimers]{muennighoff2023mtebmassivetextembedding}
Niklas Muennighoff, Nouamane Tazi, Loïc Magne, and Nils Reimers.
\newblock Mteb: Massive text embedding benchmark, 2023.
\newblock URL \url{https://arxiv.org/abs/2210.07316}.

\bibitem[Reimers \& Gurevych(2019)Reimers and Gurevych]{reimers2019sentencebertsentenceembeddingsusing}
Nils Reimers and Iryna Gurevych.
\newblock Sentence-bert: Sentence embeddings using siamese bert-networks, 2019.
\newblock URL \url{https://arxiv.org/abs/1908.10084}.

\bibitem[Son et~al.(2023)Son, Lee, Kang, and Hahm]{son2023removingnonstationaryknowledgepretrained}
Guijin Son, Hanwool Lee, Nahyeon Kang, and Moonjeong Hahm.
\newblock Removing non-stationary knowledge from pre-trained language models for entity-level sentiment classification in finance, 2023.
\newblock URL \url{https://arxiv.org/abs/2301.03136}.

\bibitem[Son et~al.(2024{\natexlab{a}})Son, Lee, Kim, Kim, Muennighoff, Choi, Park, Yoo, and Biderman]{son2024kmmlumeasuringmassivemultitask}
Guijin Son, Hanwool Lee, Sungdong Kim, Seungone Kim, Niklas Muennighoff, Taekyoon Choi, Cheonbok Park, Kang~Min Yoo, and Stella Biderman.
\newblock Kmmlu: Measuring massive multitask language understanding in korean, 2024{\natexlab{a}}.
\newblock URL \url{https://arxiv.org/abs/2402.11548}.

\bibitem[Son et~al.(2024{\natexlab{b}})Son, Lee, Kim, Kim, Lee, Yeom, Jung, Kim, and Kim]{son2024haeraebenchevaluationkorean}
Guijin Son, Hanwool Lee, Suwan Kim, Huiseo Kim, Jaecheol Lee, Je~Won Yeom, Jihyu Jung, Jung~Woo Kim, and Songseong Kim.
\newblock Hae-rae bench: Evaluation of korean knowledge in language models, 2024{\natexlab{b}}.
\newblock URL \url{https://arxiv.org/abs/2309.02706}.

\bibitem[Su et~al.(2023)Su, Shi, Kasai, Wang, Hu, Ostendorf, tau Yih, Smith, Zettlemoyer, and Yu]{su2023embeddertaskinstructionfinetunedtext}
Hongjin Su, Weijia Shi, Jungo Kasai, Yizhong Wang, Yushi Hu, Mari Ostendorf, Wen tau Yih, Noah~A. Smith, Luke Zettlemoyer, and Tao Yu.
\newblock One embedder, any task: Instruction-finetuned text embeddings, 2023.
\newblock URL \url{https://arxiv.org/abs/2212.09741}.

\bibitem[Tang \& Yang(2024)Tang and Yang]{tang2024needdomainspecificembeddingmodels}
Yixuan Tang and Yi~Yang.
\newblock Do we need domain-specific embedding models? an empirical investigation, 2024.
\newblock URL \url{https://arxiv.org/abs/2409.18511}.

\bibitem[Wang et~al.(2023)Wang, Yang, Huang, Yang, Majumder, and Wei]{wang2023improving}
Liang Wang, Nan Yang, Xiaolong Huang, Linjun Yang, Rangan Majumder, and Furu Wei.
\newblock Improving text embeddings with large language models.
\newblock \emph{arXiv preprint arXiv:2401.00368}, 2023.

\bibitem[Wang et~al.(2020)Wang, Wei, Dong, Bao, Yang, and Zhou]{wang2020minilmdeepselfattentiondistillation}
Wenhui Wang, Furu Wei, Li~Dong, Hangbo Bao, Nan Yang, and Ming Zhou.
\newblock Minilm: Deep self-attention distillation for task-agnostic compression of pre-trained transformers, 2020.
\newblock URL \url{https://arxiv.org/abs/2002.10957}.

\bibitem[Wu et~al.(2023)Wu, Irsoy, Lu, Dabravolski, Dredze, Gehrmann, Kambadur, Rosenberg, and Mann]{wu2023bloomberggptlargelanguagemodel}
Shijie Wu, Ozan Irsoy, Steven Lu, Vadim Dabravolski, Mark Dredze, Sebastian Gehrmann, Prabhanjan Kambadur, David Rosenberg, and Gideon Mann.
\newblock Bloomberggpt: A large language model for finance, 2023.
\newblock URL \url{https://arxiv.org/abs/2303.17564}.

\end{thebibliography}
\bibliographystyle{iclr2025_conference}

% --- 부록 시작 ---
\appendix
\section{Appendix}

\subsubsection*{Acknowledgments}
This research was supported by Brian Impact Foundation, a non-profit organization dedicated to the advancement of science and technology for all.

\FloatBarrier % 본문에 있는 모든 float들을 강제로 배치
\subsection{Dataset Details}

% --- Summarization ---
\begin{table}[h!]
    \centering
    \begin{tabularx}{\textwidth}{%
        >{\raggedright\arraybackslash}p{0.25\textwidth} % Dataset (20%) - 자동 줄바꿈 적용
        >{\raggedright\arraybackslash}p{0.55\textwidth} % Description (60%)
        >{\raggedright\arraybackslash}p{0.15\textwidth} % Source (20%)
    }
         \hline
         \textbf{Dataset} & \textbf{Description} & \textbf{Source} \\
         \hline
         Law-Summ-ko & Dataset for evaluating the accuracy of long text summarization in Korean legal documents using AIHub summarization data. & AIHub\footnotemark[2] \\
         News-Summ-ko & Dataset for evaluating the accuracy of long text summarization in fact-based Korean news articles. & Financial News \\
         Opinion-Summ-ko & Dataset for assessing the accuracy of long text summarization in opinion-based Korean editorials and columns using AIHub summarization data. & AIHub\footnotemark[2] \\
         FinNews-Summ-ko & Dataset for verifying the accuracy of economic news article summaries using AIHub summarization data. & AIHub\footnotemark[2] \\
         FinOpinion-Summ-ko & Dataset for evaluating the accuracy of economic editorial and column summaries using AIHub summarization data. & AIHub\footnotemark[2] \\
         \hline
    \end{tabularx}
    \caption{Summary of Summarization Datasets}
    \label{tab:my_label}
\end{table}

% --- PairClassification & Reranking & Clustering ---

\begin{table}[h!]
    \centering
    \renewcommand{\arraystretch}{1.2} % 표 행 간격 조정
    \resizebox{\textwidth}{!}{ % 표 크기 조정
    \begin{tabularx}{\textwidth}{%
        >{\raggedright\arraybackslash}p{\dimexpr 0.25\hsize\relax} % SubTask
        >{\raggedright\arraybackslash}p{\dimexpr 0.55\hsize\relax} % Description
        >{\raggedright\arraybackslash}p{\dimexpr 0.15\hsize\relax} % Source
    }
         \hline
         \textbf{Dataset} & \textbf{Description} & \textbf{Source} \\
         \hline
         HeadlineAC-PairCLS-ko & Dataset for pair classification based on collected Korean economic news headlines.  & Financial News \\
         NewsContent-PairCLS-ko & Dataset for pair classification based on collected Korean economic news articles.  & Financial News \\
         HeadlineOpinion-PairCLS-ko & Dataset for pair classification based on collected Korean economic editorial and column headlines. & Financial News \\
         FinanceFiQA-Reranking-ko & Dataset leveraging the open dataset, consisting of (query, positive) pairs along with one hard negative retrieved through hard negative mining and nine randomly selected negatives.  & Open Dataset\footnotemark[3] \\
         DARTCompany2Industry-Clustering-ko & Dataset for clustering five industry categories based on business overview sentences from DART business reports.  & DART \\
         MinDS-Clustering-ko & Dataset for clustering user requirements and intentions based on labeled user queries related to finance and insurance from the AIHub complaint Q\&A dataset.  & AIHub\footnotemark[4] \\
         \hline
    \end{tabularx}
    }
    \caption{Summary of PairClassification, Reranking, Clustering Datasets}
    \label{tab:my_label}
\end{table}

% --- classification ---
\begin{table}[h!]
    \centering
    \resizebox{\textwidth}{!}{ % ⬅️ 표 크기 조정
    \begin{tabularx}{\textwidth}{%
        >{\raggedright\arraybackslash}p{\dimexpr 0.25\hsize\relax} % SubTask
        >{\raggedright\arraybackslash}p{\dimexpr 0.55\hsize\relax} % Description
        >{\raggedright\arraybackslash}p{\dimexpr 0.15\hsize\relax} % Source
    }
         \hline
         \textbf{Dataset} & \textbf{Description} & \textbf{Source} \\
         \hline
         FinSent-CLS-ko & A dataset that classifies Korean financial news articles into Positive, Negative, or Neutral.  & Financial News \\
         ESG-CLS-ko & A dataset that selects Korean financial news articles containing the term "ESG" and classifies them into Environmental (E), Social (S), Governance (G), or Non-ESG categories.  & Financial News \\
         FOMC-CLS-ko & A dataset that filters Korean financial news articles containing the terms "interest rate," "hawkish," or "dovish" and classifies them into Hawkish, Dovish, or Neutral categories. & Financial News \\
         Industry-CLS-ko & A dataset that classifies industry analysis reports into one of ten industry sectors.  & Industry Analysis Report \\
         FinascSent-CLS-ko & A dataset that utilizes KLUE-TC and Naver Finance analysis reports to classify sentiment (Positive, Neutral, or Negative) based on specific aspects.  & Open Dataset\footnotemark[5]  \\
         FinancialMMLU-CLS-ko & An open dataset built on public websites and Wikipedia, designed for multiple-choice question answering in the financial domain, where a question and answer choices are provided. & Open Dataset\footnotemark[6] \\
         FinancialBQA-CLS-ko & An open dataset built on AI Hub data, designed for binary question answering (Binary QA), where a given context is read and answered with Yes or No.  & Open Dataset\footnotemark[7] \\
         FInancialMCQA-CLS-ko & An open dataset built on AI Hub data, designed as a multiple-choice QA subset, where a given context is read and answered by selecting one of the provided choices. & Open Dataset\footnotemark[7] \\
         FinNewsBQA-CLS-ko & A financial news machine reading comprehension dataset from AI Hub, where economic news texts and queries are provided, and the queries are answered with Yes or No in a binary QA format.  & AIHub\footnotemark[8] \\
         \hline
    \end{tabularx}
    } % ⬅️ resizebox 닫기
    \caption{Summary of Classification Datasets}
    \label{tab:my_label}
\end{table}

% --- STS & Retrieval ---
\begin{table}[h!]
    \centering
    \begin{tabularx}{\textwidth}{%
        >{\raggedright\arraybackslash}p{0.25\textwidth} % Dataset (20%) - 자동 줄바꿈 적용
        >{\raggedright\arraybackslash}p{0.55\textwidth} % Description (60%)
        >{\raggedright\arraybackslash}p{0.15\textwidth} % Source (20%)
    }
         \hline
         \textbf{Dataset} & \textbf{Description} & \textbf{Source} \\
         \hline
         FinSTS-ko & STS dataset for detecting subtle semantic shifts in financial narratives, constructed using sentence pairs collected from financial news. & Financial News \\
         TATQA-Retrieval-ko & Textual and tabular QA dataset constructed using information extracted from financial reports. & Financial Report \\
         FinNews-Retrieval-ko & Retrieval dataset constructed using Korean financial news articles. & Financial News \\
         BokFinDict-Retrieval-ko & Retrieval dataset tailored for the Korean financial domain, based on the Bank of Korea financial dictionary. & BoK(Bank of Korea)Financial Dictionary \\
         FSSFinDict-Retrieval-ko& Retrieval dataset tailored for the Korean financial domain, based on the Financial Supervisory Service (FSS) financial dictionary. & FSS(Financial Supervisory Service)Financial Dictionary \\
         FinMarketReport-Retrieval-ko & Retrieval dataset for the stock market, constructed using financial reports. & Financial Report \\
         \hline
    \end{tabularx}
    \caption{Summary of STS, Retrieval Datasets}
    \label{tab:my_label}
\end{table}

\footnotetext[2]{\url{https://aihub.or.kr/aihubdata/data/view.do?currMenu=115&topMenu=100&aihubDataSe=realm&dataSetSn=97}}
\footnotetext[3]{\url{https://huggingface.co/datasets/BCCard/BCCard-Finance-Kor-QnA}}
\footnotetext[4]{\url{https://aihub.or.kr/aihubdata/data/view.do?currMenu=115&topMenu=100&aihubDataSe=realm&dataSetSn=98}}
\footnotetext[5]{\url{https://huggingface.co/datasets/amphora/korfin-asc}}
\footnotetext[6]{\url{https://huggingface.co/datasets/allganize/financial-mmlu-ko}}
\footnotetext[7]{\url{https://huggingface.co/datasets/FINNUMBER/QA_Instruction}}
\footnotetext[8]{\url{https://www.aihub.or.kr/aihubdata/data/view.do?currMenu=115&topMenu=100&dataSetSn=577}}

\FloatBarrier

\subsection{Experiment Results}
% Table 1: Classification 결과
\begin{table}[H]
    \centering
    \resizebox{\textwidth}{!}{%
    \begin{tabular}{cccccl}
         \hline
         \textbf{Task} & \textbf{Model} & \textbf{Metric} & \textbf{FinMTEB Score} & \textbf{KorFinMTEB Score} & \textbf{Diff} \\
         \hline
         \multirow{8}{*}{FOMCClassification} 
         & bge-en-icl                & accuracy & 0.870 & 0.690 & \textcolor{red}{$\blacktriangle$} 0.180 \\
         & gte-Qwen2-1.5B-instruct   & -- & 0.545 & 0.705 & \textcolor{blue}{$\blacktriangledown$} -0.160 \\
         & e5-mistral-7b-instruct    & -- & 0.865 & 0.700 & \textcolor{red}{$\blacktriangle$} 0.165 \\
         & bge-large-en-v1.5         & -- & 0.485 & 0.680 & \textcolor{blue}{$\blacktriangledown$} -0.195 \\
         & text-embedding-3-small    & -- & 0.730 & 0.580 & \textcolor{red}{$\blacktriangle$} 0.150 \\
         & all-MiniLM-L12-v2         & -- & 0.480 & 0.640 & \textcolor{blue}{$\blacktriangledown$} -0.160 \\
         & instructor-base           & -- & 0.450 & 0.640 & \textcolor{blue}{$\blacktriangledown$} -0.190 \\
         & kure-v1                   & -- & 0.585   & 0.725 & \textcolor{blue}{$\blacktriangledown$} -0.140 \\
         \hline
         \multirow{8}{*}{ESGClassification} 
         & bge-en-icl                & accuracy & 0.620 & 0.750 & \textcolor{blue}{$\blacktriangledown$} -0.130 \\
         & gte-Qwen2-1.5B-instruct   & -- & 0.865 & 0.705 & \textcolor{red}{$\blacktriangle$} 0.160 \\
         & e5-mistral-7b-instruct    & -- & 0.610 & 0.765 & \textcolor{blue}{$\blacktriangledown$} -0.155 \\
         & bge-large-en-v1.5         & -- & 0.645 & 0.565 & \textcolor{red}{$\blacktriangle$} 0.080 \\
         & text-embedding-3-small    & -- & 0.810 & 0.710 & \textcolor{red}{$\blacktriangle$} 0.100 \\
         & all-MiniLM-L12-v2         & -- & 0.580 & 0.565 & \textcolor{red}{$\blacktriangle$} 0.015 \\
         & instructor-base           & -- & 0.500 & 0.555 & \textcolor{blue}{$\blacktriangledown$} -0.055 \\
         & kure-v1                   & -- & 0.850   & 0.690   & \textcolor{red}{$\blacktriangle$} 0.160 \\
         \hline
         \multirow{8}{*}{FinNewsClassification} 
         & bge-en-icl                & accuracy & 0.775 & 0.730 & \textcolor{red}{$\blacktriangle$} 0.045 \\
         & gte-Qwen2-1.5B-instruct   & -- & 0.670 & 0.610 & \textcolor{red}{$\blacktriangle$} 0.060 \\
         & e5-mistral-7b-instruct    & -- & 0.755 & 0.725 & \textcolor{red}{$\blacktriangle$} 0.030 \\
         & bge-large-en-v1.5         & -- & 0.510 & 0.445 & \textcolor{red}{$\blacktriangle$} 0.065 \\
         & text-embedding-3-small    & -- & 0.665 & 0.570 & \textcolor{red}{$\blacktriangle$} 0.095 \\
         & all-MiniLM-L12-v2         & -- & 0.545 & 0.445 & \textcolor{red}{$\blacktriangle$} 0.100 \\
         & instructor-base           & -- & 0.470 & 0.405 & \textcolor{red}{$\blacktriangle$} 0.065 \\
         & kure-v1                   & -- & 0.710 & 0.715 & \textcolor{blue}{$\blacktriangledown$} -0.005 \\
         \hline
    \end{tabular}
    }
    \caption{Result of FinMTEB-KorFinMTEB Classification Task.}
    \label{tab:table_1}
\end{table}

% \FloatBarrier

% Table 2: FinSTS 결과
\begin{table}[H]
    \centering
    \resizebox{\textwidth}{!}{%
    \begin{tabular}{cccccl}
         \hline
         \textbf{Task} & \textbf{Model} & \textbf{Metric} & \textbf{*FinMTEB Score} & \textbf{KorFinMTEB Score} & \textbf{Diff} \\
         \hline
         \multirow{8}{*}{FinSTS} 
         & bge-en-icl                & cosine\_spearman & 0.135 & 0.030  & \textcolor{red}{$\blacktriangle$} 0.096\\
         & gte-Qwen2-1.5B-instruct   & --               & 0.095 & -0.085  & \textcolor{red}{$\blacktriangle$} 0.180\\
         & e5-mistral-7b-instruct    & --               & 0.142 & 0.0250  & \textcolor{red}{$\blacktriangle$} 0.116\\
         & bge-large-en-v1.5         & --               & 0.030 & -0.059 & \textcolor{red}{$\blacktriangle$} 0.089\\
         & text-embedding-3-small    & --               & 0.158 & -0.025 & \textcolor{red}{$\blacktriangle$} 0.183\\
         & all-MiniLM-L12-v2         & --               & 0.050 & -0.003 & \textcolor{red}{$\blacktriangle$} 0.054\\
         & instructor-base           & --               & -0.023 & 0.049  & \textcolor{blue}{$\blacktriangledown$} -0.073\\
         & kure-v1                   & --               & 0.225   & 0.078   & \textcolor{red}{$\blacktriangle$} 0.147\\
         \hline
    \end{tabular}
    }
    \caption{Result of FinMTEB- KorFinMTEB  STS Task}
    \label{tab:table_3}
\end{table}

% \FloatBarrier

% Table 3: PairClassification, FiQA2018Reranking, Clustering 결과
\begin{table}[H]
    \centering
    \resizebox{\textwidth}{!}{%
    \begin{tabular}{cccccl}
         \hline
         \textbf{Task} & \textbf{Model} & \textbf{Metric} & \textbf{FinMTEB Score} & \textbf{KorFinMTEB Score} & \textbf{Diff} \\
         \hline
         \multirow{8}{*}{HeadlineACPairClassification} 
         & bge-en-icl                & average precision & 0.761 & 0.436 & \textcolor{red}{$\blacktriangle$} 0.325\\
         & gte-Qwen2-1.5B-instruct   & --                & 0.763 & 0.429 & \textcolor{red}{$\blacktriangle$} 0.334\\
         & e5-mistral-7b-instruct    & --                & 0.744 & 0.409 & \textcolor{red}{$\blacktriangle$} 0.335\\
         & bge-large-en-v1.5         & --                & 0.751 & 0.437 & \textcolor{red}{$\blacktriangle$} 0.314\\
         & text-embedding-3-small    & --                & 0.763 & 0.442 & \textcolor{red}{$\blacktriangle$} 0.321\\
         & all-MiniLM-L12-v2         & --                & 0.725 & 0.459 & \textcolor{red}{$\blacktriangle$} 0.266\\
         & instructor-base           & --                & 0.707 & 0.252 & \textcolor{red}{$\blacktriangle$} 0.455\\
         & kure-v1                   & --                & 0.755 & 0.669 & \textcolor{red}{$\blacktriangle$} 0.086\\
         \hline
         \multirow{8}{*}{FiQA2018Reranking} 
         & bge-en-icl                & mAP               & 0.961 & 0.436 & \textcolor{red}{$\blacktriangle$} 0.525\\
         & gte-Qwen2-1.5B-instruct   & --                & 0.722 & 0.124 & \textcolor{red}{$\blacktriangle$} 0.598\\
         & e5-mistral-7b-instruct    & --                & 0.732 & 0.364 & \textcolor{red}{$\blacktriangle$} 0.368\\
         & bge-large-en-v1.5         & --                & 0.751 & 0.437 & \textcolor{red}{$\blacktriangle$} 0.314\\
         & text-embedding-3-small    & --                & 0.760 & 0.440 & \textcolor{red}{$\blacktriangle$} 0.320\\
         & all-MiniLM-L12-v2         & --                & 0.725 & 0.459 & \textcolor{red}{$\blacktriangle$} 0.266\\
         & instructor-base           & --                & 0.707 & 0.252 & \textcolor{red}{$\blacktriangle$} 0.455\\
         & kure-v1                   & --                & 0.755 & 0.669 & \textcolor{red}{$\blacktriangle$} 0.086\\
         \hline
         \multirow{7}{*}{dart\_company2industry\_clustering} 
         & bge-en-icl                & v\_measure        & 0.725 & 0.248 & \textcolor{red}{$\blacktriangle$} 0.477\\
         & gte-Qwen2-1.5B-instruct   & --                & 0.729 & 0.345 & \textcolor{red}{$\blacktriangle$} 0.384\\
         & e5-mistral-7b-instruct    & --                & 0.730 & 0.369 & \textcolor{red}{$\blacktriangle$} 0.361\\
         & bge-large-en-v1.5         & --                & 0.203 & 0.160 & \textcolor{red}{$\blacktriangle$} 0.043\\
         & text-embedding-3-small    & --                & 0.659 & 0.245 & \textcolor{red}{$\blacktriangle$} 0.414\\
         & all-MiniLM-L12-v2         & --                & 0.081 & 0.045 & \textcolor{red}{$\blacktriangle$} 0.036\\
         & instructor-base           & --                & 0.109 & 0.060 & \textcolor{red}{$\blacktriangle$} 0.049\\
         & kure-v1                   & --                & 0.642 & 0.321 & \textcolor{red}{$\blacktriangle$} 0.321\\
         \hline
    \end{tabular}
    }
    \caption{Result of FinMTEB-KorFinMTEB PairClassification, FiQA2018Reranking, Clustering Task }
    \label{tab:table_2}
\end{table}

% Table 4: Retrieval 결과
\begin{table}[H]
    \centering
    \resizebox{\textwidth}{!}{%
    \begin{tabular}{cccccl}
         \hline
         \textbf{Task} & \textbf{Model} & \textbf{Metric} & \textbf{FinMTEB Score} & \textbf{KorFinMTEB Score} & \textbf{Diff} \\
         \hline
         \multirow{8}{*}{TAT QA Retrieval} 
         & bge-en-icl                & ndcg@10 & 0.408 & 0.300 & \textcolor{red}{$\blacktriangle$} 0.108\\
         & gte-Qwen2-1.5B-instruct   & --                & 0.406 & 0.801 & \textcolor{blue}{$\blacktriangledown$} -0.395\\
         & e5-mistral-7b-instruct    & --                & 0.354 & 0.816 & \textcolor{blue}{$\blacktriangledown$} -0.462\\
         & bge-large-en-v1.5         & --                & 0.360 & 0.288 & \textcolor{red}{$\blacktriangle$} 0.072\\
         & text-embedding-3-small    & --                & 0.273 & 0.793 & \textcolor{blue}{$\blacktriangledown$} -0.520\\
         & all-MiniLM-L12-v2         & --                & 0.282 & 0.202 & \textcolor{red}{$\blacktriangle$} 0.080\\
         & instructor-base           & --                & 0.360 & 0.163 & \textcolor{red}{$\blacktriangle$} 0.197\\
         & kure-v1                   & --                & 0.109 & 0.802 & \textcolor{blue}{$\blacktriangledown$} -0.693\\
         \hline
         \multirow{8}{*}{GoldmanEncRetrieval (vs. FssDict)} 
         & bge-en-icl                & ndcg@10               & 0.400 & 0.500 & \textcolor{blue}{$\blacktriangledown$} -0.100\\
         & gte-Qwen2-1.5B-instruct   & --                & 0.446 & 0.719 & \textcolor{blue}{$\blacktriangledown$} -0.273\\
         & e5-mistral-7b-instruct    & --                & 0.448 & 0.708 & \textcolor{blue}{$\blacktriangledown$} -0.260\\
         & bge-large-en-v1.5         & --                & 0.542 & 0.760 & \textcolor{blue}{$\blacktriangledown$} -0.218\\
         & text-embedding-3-small    & --                & 0.257 & 0.407 & \textcolor{blue}{$\blacktriangledown$} -0.150\\
         & all-MiniLM-L12-v2         & --                & 0.522 & 0.600 & \textcolor{blue}{$\blacktriangledown$} -0.078\\
         & instructor-base           & --                & 0.344 & 0.360 & \textcolor{blue}{$\blacktriangledown$} -0.016\\
         & kure-v1                   & --                & 0.334 & 0.606 & \textcolor{blue}{$\blacktriangledown$} -0.272\\
         \hline
         \multirow{7}{*}{GoldmanEncRetrieval (vs. BokDict)} 
         & bge-en-icl                & ndcg@10        & 0.410 & 0.350 & \textcolor{red}{$\blacktriangle$} 0.060\\
         & gte-Qwen2-1.5B-instruct   & --                & 0.446 & 0.742 & \textcolor{blue}{$\blacktriangledown$} -0.296\\
         & e5-mistral-7b-instruct    & --                & 0.448 & 0.737 & \textcolor{blue}{$\blacktriangledown$} -0.289\\
         & bge-large-en-v1.5         & --                & 0.542 & 0.790 & \textcolor{blue}{$\blacktriangledown$} -0.248\\
         & text-embedding-3-small    & --                & 0.257 & 0.405 & \textcolor{blue}{$\blacktriangledown$} -0.148\\
         & all-MiniLM-L12-v2         & --                & 0.522 & 0.530 & \textcolor{blue}{$\blacktriangledown$} -0.008\\
         & instructor-base           & --                & 0.344 & 0.370 & \textcolor{blue}{$\blacktriangledown$} -0.026\\
         & kure-v1                   & --                & 0.334 & 0.582 & \textcolor{blue}{$\blacktriangledown$} -0.248\\
         \hline
    \end{tabular}
    }
    \caption{Result of FinMTEB-KorFinMTEB Retrieval Task }
    \label{tab:table_2}
\end{table}

\FloatBarrier

\end{document}